\documentclass{article}

\usepackage{microtype}
\usepackage{graphicx}
\usepackage{subfigure}
\usepackage{booktabs} 
\usepackage{hyperref}

\usepackage[accepted]{icml2020}

\icmltitlerunning{Learning Representations that Support Extrapolation}

\begin{document}

\twocolumn[
\icmltitle{Learning Representations that Support Extrapolation}

\begin{icmlauthorlist}
\icmlauthor{Taylor W. Webb}{ucla}
\icmlauthor{Zachary Dulberg}{pni}
\icmlauthor{Steven M. Frankland}{pni}
\icmlauthor{Alexander A. Petrov}{osu}
\icmlauthor{Randall C. O'Reilly}{ucd}
\icmlauthor{Jonathan D. Cohen}{pni}
\end{icmlauthorlist}

\icmlaffiliation{ucla}{Department of Psychology, University of California Los Angeles, Los Angeles, CA}
\icmlaffiliation{pni}{Princeton Neuroscience Institute, Princeton, NJ}
\icmlaffiliation{osu}{Department of Psychology, The Ohio State University, Columbus, OH}
\icmlaffiliation{ucd}{Department of Psychology, University of California Davis, Davis, CA}

\icmlcorrespondingauthor{Taylor Webb}{taylor.w.webb@gmail.com}

\icmlkeywords{extrapolation, analogy, out-of-domain generalization, normalization}

\vskip 0.3in
]

\printAffiliationsAndNotice{}  

\begin{abstract}
Extrapolation -- the ability to make inferences that go beyond the scope of one's experiences -- is a hallmark of human intelligence. By contrast, the generalization exhibited by contemporary neural network algorithms is largely limited to interpolation between data points in their training corpora. In this paper, we consider the challenge of learning representations that support extrapolation. We introduce a novel visual analogy benchmark that allows the graded evaluation of extrapolation as a function of distance from the convex domain defined by the training data. We also introduce a simple technique, context normalization, that encourages representations that emphasize the relations between objects. We find that this technique enables a significant improvement in the ability to extrapolate, considerably outperforming a number of competitive techniques.

\end{abstract}

\section{Introduction}

The notion of interpolation is built into the assumptions underlying most approaches to generalization in machine learning, in which it is typically assumed that training and test samples are drawn from the same distribution. There is a widely shared view that human reasoning involves something more than this, that human reasoning involves the ability to \textit{extrapolate} \cite{lake2017building, marcus2001algebraic}. In particular, advocates of this view sometimes point to analogical reasoning as a clear example of this capacity, in which a reasoner extrapolates from knowledge in one domain to make inferences in a different, often less familiar, domain, based on common structure between the two \cite{gentner1983structure, holyoak2012analogy}.

What are the prospects for capturing the capacity for extrapolation in neural network algorithms? Recent results have begun to address this question. In general, these results point to the conclusion that generalization in neural networks, even in relatively sophisticated domains such as relational or mathematical reasoning, is primarily limited to interpolation between data points within the convex hull (i.e. boundaries defined by the extremes) of the training set \cite{lake2017generalization, santoro2018measuring, hill2019learning, saxton2019analysing}.

It is worth considering how extrapolation is possible at all. Consider, for example, theoretical physics, a spectacularly successful paradigm of extrapolation. Physical laws discovered on the basis of terrestrial observations make precise quantitative predictions about phenomena in distant galaxies. This is possible because physical laws are characterized by certain symmetries -- that is, they are invariant with respect to a group of transformations such as translation and rotation in space, translation in time, etc. \cite{feynman1966symmetry}. As Feynman puts it, ``nature uses only the longest threads to weave her patterns, so each small piece of her fabric reveals the organization of the entire tapestry" \cite{feynman1967character}.

In this work, we exploit the fact that many domains can be characterized by such symmetries, and test the idea that extrapolation can be enabled by encouraging the learning of representations that reflect these symmetries. To do so, we introduce \textit{context normalization}, a simple inductive bias in which normalization is applied over a task-relevant temporal window. This technique preserves local relational information (e.g. the size of one object relative to another), while introducing both scale and translation invariance over the broader scope of the learned representational space. We hypothesized that the application of context normalization would improve the ability of neural networks to extrapolate. Critically, when trained end-to-end, we expect the presence of context normalization to impose constraints on both upstream (computed prior to a layer with normalization) and downstream (computed post-normalization) representations, promoting the acquisition of abstract representations that reflect task-relevant symmetries.

In an effort to aid the systematic evaluation of extrapolation in neural networks, we also introduce a novel benchmark, the Visual Analogy Extrapolation Challenge (VAEC). This dataset has two major advantages relative to other benchmarks designed to evaluate extrapolation \cite{santoro2018measuring,hill2019learning, saxton2019analysing}. First, VAEC contains generalization regimes that assess both translation and scale invariance with respect to the underlying task space. Second, in each regime, VAEC includes a progressive series of evaluation modes, in which test data lie increasingly far away from the convex hull defined by the training data, allowing the graded evaluation of extrapolation as a function of distance from the training domain. We evaluate context normalization, in addition to a number of competitive alternative techniques, on the VAEC dataset, the visual analogy dataset from Hill et al. \yrcite{hill2019learning}, and a dynamic object prediction task. We find that context normalization yields a considerable improvement in the ability to extrapolate in each of these task domains. 

\section{Task Setup}

\subsection{VAEC Dataset}

\begin{figure}[ht]
\vskip 0.2in
\begin{center}
\centerline{\includegraphics[width=\columnwidth]{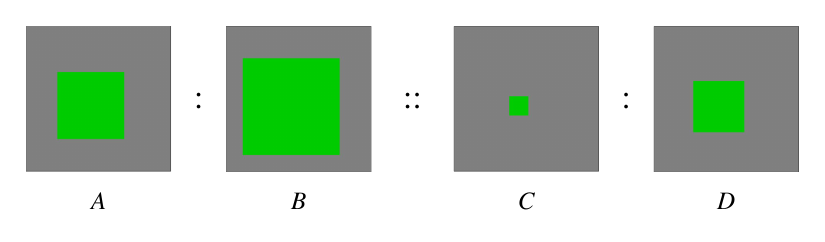}}
\caption{Example visual analogy problem.}
\label{analogy-task}
\end{center}
\vskip -0.2in
\end{figure}

The VAEC dataset consists of four-term visual analogy problems, constructed from objects that vary in brightness, size, and location along the X and Y axes. Each object is rendered as an RGB image of size $128\times128\times3$, with each color channel scaled to produce a minimum possible value of $0$ and maximum possible value of $1$. Each image consists of a green square on a gray background ($R = 0.5$, $G = 0.5$, $B = 0.5$). Each of the four dimensions of variation consists of a linear range tiled by $42$ discrete levels, with brightness spanning the range $G \in[0.4\cdots1.0]$, size spanning the range $width \in[3\cdots85]$, and location spanning the range $center \in[43\cdots84]$ along both X and Y axes. 

The dataset contains proportional analogy problems of the form $A:B::C:D$, in which all four terms of a given problem vary along one stimulus dimension, and in which both the distance and direction along that dimension are the same for $A$ and $B$ as they are for $C$ and $D$ (Figure~\ref{analogy-task}). Each analogy problem also contains a set of $6$ foil objects $F_{1}$ through $F_{6}$, each of which take the same values as the terms of the analogy ($A$, $B$, $C$, and $D$) along the irrelevant dimensions of a problem, but take a different value than $D$ along the relevant dimension. The task is to select $D$ from a set of multiple choices consisting of $D$ and $F_{1}$ through $F_{6}$.

\begin{figure}
\vskip 0.2in
\begin{center}
\centerline{\includegraphics[width=\columnwidth]{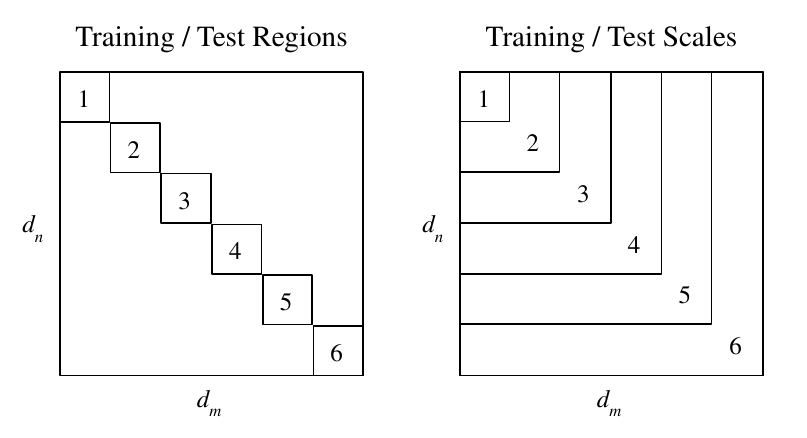}}
\caption{Illustration of the Translation Extrapolation regime, divided into Regions $1$-$6$, and Scale Extrapolation regime, divided into Scales $1$-$6$. The underlying visual object space is illustrated here in two dimensions ($d_{n}$ and $d_{m}$), though the actual space consists of four dimensions (size, brightness, and 2d location).}
\label{gen_regimes}
\end{center}
\vskip -0.2in
\end{figure}

\subsubsection{Translation Extrapolation Regime}

The VAEC dataset contains two generalization regimes, each requiring a distinct form of extrapolation. The Translation Extrapolation regime tests for invariance to translation along the underlying dimensions of the visual object space (size, brightness, and location), by dividing the space into six regions along the diagonal, each with size $7\times7\times7\times7$. These are referred to as Regions $1$-$6$ (Figure~\ref{gen_regimes}), where Region $1$ consists of small, dim objects located in the upper left portion of space, and Region $6$ consists of large, bright objects located in the lower right portion of space. This allows the graded evaluation of extrapolation to a series of increasingly remote test domains.

We note that, although the sources of variation (size, brightness, and location) are correlated in terms of how training and test regions are defined, they are not correlated within an individual analogy problem. Each individual analogy problem only includes a single source of variation. The source of variation in each problem is included as an annotation in the dataset, and can be used to analyze whether and how performance differs across these dimensions. We include such an analysis in the Supplementary Material.

For each region, analogy problems are subsampled from the set of all valid analogies, resulting in a dataset with $19,040$ analogy problems per region, approximately $10\%$ of all valid analogies within a region. In our work, we train networks on analogies from Region $1$, and test on analogies from Regions $2$ through $6$, though we note that other configurations are possible with the dataset. Critically, test samples involve not only novel objects, but objects that fall completely outside the range of values observed during training.

\subsubsection{Scale Extrapolation Regime}

The VAEC dataset also contains a second generalization regime, the Scale Extrapolation regime, that tests for invariance to the scale of differences between visual objects. This regime includes six evaluation modes, referred to as Scale $1$ through Scale $6$. Scale $1$ contains objects sampled from levels $1$-$7$ along each visual object dimension (identical to values used in Region $1$ of the Translation Extrapolation regime). In scales $2$ through $6$, these values are multiplied by a scalar ranging from $2$ to $6$. Just as with the Translation Extrapolation regime, analogy problems in the Scale Extrapolation regime are: a) subsampled from the set of all valid analogies at each scale, resulting in $19,040$ analogy problems per scale; and b) use test samples outside the range of values observed during training.

\subsection{Visual Analogy Dataset}

We also evaluated context normalization on the extrapolation regime from the visual analogy dataset in Hill et al. \yrcite{hill2019learning}. Inspired by Raven's Progressive Matrices \cite{raven1941standardization}, this dataset consists of $2 \times 3$ matrices, in which a rule must be inferred from the images in the first row (the \textit{source}) and then applied to the images in the second row (the \textit{target}). Although the extrapolation regime in this dataset is in some ways easier than the VAEC dataset, because it does not require extrapolation as far away from the training domain, it is also more challenging in some ways, because it involves distracting, task-irrelevant variation, it involves cross-domain analogies (e.g. mapping a change in brightness to a change in size), and because each image typically involves multiple objects. 

\subsection{Dynamic Object Prediction Task}

In order to evaluate the generality of our proposed context normalization method, we also evaluated context normalization on a dynamic object prediction task. Specifically, we created a task containing a sequence of  $T$ images, $\textbf{x}_{1}$\ldots$\textbf{x}_{T}$, depicting a smoothly changing object, requiring the prediction of the image $\textbf{x}_{t}$ given images $\textbf{x}_{1}$\ldots$\textbf{x}_{t-1}$, for each time step $t$ in the sequence. We used grayscale images of size $64\times64$, each containing a white square on a black background. Over the course of a sequence, the location and size of the square changed smoothly. Location varied within the range $center \in[16\cdots48]$ along the X and Y axes, and size varied within the range $width \in[3\cdots31]$.

\begin{figure}
\vskip 0.2in
\begin{center}
\centerline{\includegraphics[width=0.7\columnwidth]{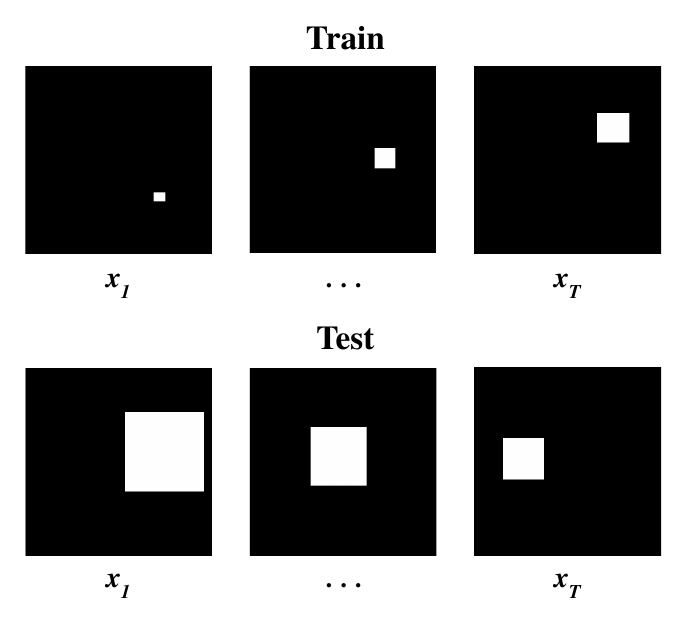}}
\caption{Example sequences from the dynamic object prediction task. Note the difference in object size between the train and test examples.}
\label{moving_objects}
\end{center}
\vskip -0.2in
\end{figure}

For each sequence, we uniformly sampled start and end values for object size and location and generated a sequence by linearly interpolating between these values. We used sequences with length $T = 20$. To evaluate extrapolation, we stipulated that training would be performed only on objects with sizes from the range $width \in[3\cdots13]$, and evaluation would be performed on objects with sizes from the range $width \in[13\cdots31]$. This task thus required extrapolation from training on one set of objects to testing on objects that were an average of nearly three times as large (Figure~\ref{moving_objects}).

\section{Approach}

Our central proposal for improving extrapolation is to normalize representations with respect to a task-relevant temporal window, preserving the relations between these representations, but discarding information about their absolute magnitude. We first formalize the proposed normalization approach, then describe how we apply this approach to different task domains.

\subsection{Context Normalization}

Given a batch with $N$ sequences, in which each sequence contains $T$ time points, and in which a vector with $H$ dimensions is presented at each time point, we refer to the activity in the $i$th sequence, at the $j$th time point, in the $k$th dimension as $z_{i j k}$. We define the corresponding context-normalized activity $CN(z_{i j k})$ as:

\begin{equation}
\mu_{i k}=\frac{1}{T} \sum_{j=1}^{T}\left(z_{i j k}\right)
\end{equation}

\begin{equation}
\sigma_{i k}= \sqrt{\frac{1}{T} \sum_{j=1}^{T}\left(z_{i j k}-\mu_{i k}\right)^{2} + \epsilon} 
\end{equation}

\begin{equation}
CN(z_{i j k})=\gamma_{k}(\frac{z_{i j k}-\mu_{i k}}{\sigma_{i k}}) + \beta_{k}
\end{equation}

where $\epsilon$ is a small constant to avoid division by zero, and $\gamma$ and $\beta$ are learned gain and shift parameters (all initialized to one and zero respectively). This approach is similar to batch normalization \cite{ioffe2015batch} except that it normalizes over the temporal dimension instead of the batch dimension. In our experiments, we also evaluate a range of other normalization techniques, including batch normalization.

In the proposed approach, the context over which normalization is applied can be tailored based on knowledge of the structure of the problem. In the visual analogy dataset from Hill et al. \yrcite{hill2019learning}, analogy problems in the test set require extrapolation from familiar values in the source domain to novel values in the target domain. For this dataset, we therefore implement context normalization by treating the source and target as separate contexts over which to normalize (\textit{Context norm. (source/target)} in Table~\ref{visual_analogy_results}), and compare to a version that treats the entire analogy as a single context (\textit{Context norm. (entire analogy)} in Table~\ref{visual_analogy_results}). For all other datasets, context normalization is applied over an entire analogy problem or sequence.

\subsection{Analogy Scoring Model}

To solve analogy problems in both the VAEC dataset and the visual analogy dataset from Hill et al. \yrcite{hill2019learning}, we employ an approach also proposed by Hill et al. \yrcite{hill2019learning}, treating analogy as a scoring problem. For each analogy problem, our network is presented with multiple candidate analogies, consisting of the objects $A$, $B$, $C$, and a candidate answer, drawn from the set containing $D$ and foil objects $F_{1}$ through $F_{6}$. The network produces a score for each candidate analogy, these scores are passed through a softmax function, and the network is trained to maximize the probability that $D$ is the correct answer.

The network consists of a feedforward encoder that generates a vector embedding $\textbf{z}$ for each image, a recurrent network that sequentially processes the vector embedding of each image in a candidate analogy, and a linear output layer (detailed in \ref{analogy_hyperparams}). In our experiments, we apply context normalization, along with a range of other techniques, to the vector embeddings (\textbf{z}) before passing them into the recurrent network.

\subsection{Dynamic Object Prediction Model}

To address the dynamic object prediction task, we employ an approach that combines an autoencoder and a recurrent network (detailed in \ref{dynamic_hyperparams}). First, we train an autoencoder to generate a low-dimensional embedding $\textbf{z}$ given an image $\textbf{x}$. Then, for each sequence of images $\textbf{x}_{1}$\ldots$\textbf{x}_{T}$, we obtain the corresponding low-dimensional embeddings $\textbf{z}_{1}$\ldots$\textbf{z}_{T}$. Finally, we train a recurrent network to predict $\textbf{z}_{t}$ given $\textbf{z}_{1}$\ldots$\textbf{z}_{t-1}$. The combined system is capable of making predictions in image space given an input sequence of images, and can be fine-tuned end-to-end.

In our experiments, we apply context normalization to the embeddings $\textbf{z}_{1}$\ldots$\textbf{z}_{t-1}$ before passing them to the recurrent network to make predictions. Then, before passing the predictions $\hat{\textbf{z}}_{2}$\ldots$\hat{\textbf{z}}_{T}$ to the decoder to generate predictions in image space, we invert the transformation imposed by normalization. We do this because normalization removes information about the absolute magnitude of the object that is necessary to accurately render an image. Specifically, given the prediction $\hat{z}_{i j k}$ (the activity in the $i$th sequence, at the $j$th time point, in the $k$th dimension), we define the de-normalized version $CN^{-1}(\hat{z}_{i j k})$ as:

\begin{equation}
CN^{-1}(\hat{z}_{i j k})=\sigma_{i k} \cdot \hat{z}_{i j k} + \mu_{i k}
\end{equation}

When testing other normalization procedures on the dynamic object prediction task, we similarly invert the normalization procedure (scaling by $\sigma$ and adding $\mu$) before passing $\hat{\textbf{z}}$ to the decoder.

\section{Experiments}

\subsection{Analogy Architecture and Training Procedure}
\label{analogy_hyperparams}

For the analogy scoring model used on the VAEC dataset, the encoder architecture consisted of $4$ convolutional layers, each with $32$ kernels of size $4\times4$, and a stride of $2$ (no max-pooling), resulting in a feature map of size $8\times8\times32$. This was followed by $2$ fully-connected layers, with $256$ units per layer. ReLU nonlinearities were used in all layers of the encoder. The image embedding $\textbf{z}$ was then generated with a linear layer consisting of 256 units. 

We applied either context normalization, or one out of a number of other normalization techniques (detailed in \ref{baselines_section}), to these embeddings, amd then passed a sequence consisting of the embeddings for $A$, $B$, $C$, and the candidate answer to an LSTM with a single hidden layer of $256$ units. The final hidden state of the LSTM was then passed through a linear layer to generate a score for the candidate answer. This process was repeated for each candidate answer (using the same encoder parameters for each image, and the same recurrent and output layer parameters for each sequence, reinitializing the recurrent state at the beginning of each sequence), and the resulting scores were passed through a softmax function to generate a probability distribution over the candidate answers.

We trained networks to maximize the probability that $D$ was the correct answer using a cross entropy loss. Each network was trained for $10,000$ iterations, with a batch size of $32$, using the ADAM optimizer \cite{kingma2014adam} with a learning rate of $5e^-4$ (except as otherwise noted in \ref{baselines_section}). All weights were initialized using Xavier uniform initialization \cite{glorot2010understanding}, and all biases were initialized to zero. All simulations on the VAEC dataset were performed using TensorFlow \cite{abadi2016tensorflow}.

For the analogy scoring model used on the visual analogy dataset from Hill et al. \yrcite{hill2019learning}, we used an architecture and training procedure modeled as closely as possible on the original paper. We describe these in detail in the Supplementary Material. All simulations for this dataset were performed using PyTorch \cite{paszke2017automatic}.

\subsection{Comparison with Other Normalization Techniques}
\label{baselines_section}

We considered a wide range of techniques as alternatives to context normalization. First, we compared to \textit{batch normalization} \cite{ioffe2015batch}, in which normalization statistics are computed over the batch dimension. Just as with context normalization, we applied batch normalization to the embedding vector $\textbf{z}$ for each image in a sequence. 

In our default implementation, we evaluate performance on the test set in batches with the same size ($N = 32$) as during training, and compute batch normalization statistics online directly from these test batches. We did this to give batch normalization the best possible chance of extrapolating to domains with statistics that are different than the training set, but we note that, in the standard approach to batch normalization, normalization statistics during test are computed from the \textit{training set}, to prevent the need to have batches during evaluation. We therefore also compared to a model that included batch normalization with statistics during evaluation computed from the entire training set.

We also compared to \textit{layer normalization} \cite{ba2016layer}, in which normalization statistics are computed over the units in a hidden layer. Given that layer normalization has been proposed specifically in the context of recurrent networks, we evaluated two versions, one in which normalization was applied to the hidden layer of the LSTM, and one in which it was applied to outputs of the feedforward encoder ($\textbf{z}$). We found that we had to train the networks with layer normalization significantly longer ($500,000$ training iterations) to achieve a comparable degree of convergence on the training set.

In our experiments, batch normalization statistics were computed over a larger sample ($N = 32$) than context normalization statistics ($T = 4$). To determine whether this factor affected performance, we compared to \textit{sub-batch normalization}, in which normalization statistics were computed over sub-batches of size $4$ (though batch size was still $32$). Thus, sub-batch normalization was performed over the same dimension as batch normalization, but with sample sizes comparable to context normalization.

We also compared to a combined form of context- and batch-normalization, in which normalization statistics were computed over both the temporal and batch dimensions (similar to the `sequence-wise normalization' proposed by Laurent et al. \yrcite{laurent2016batch}).

Our proposed approach to context normalization is aligned with the temporal structure of our task, in that normalization statistics are computed over the $4$ terms of a candidate analogy. To determine the importance of this alignment, we compared to two control conditions, each of which involved first concatenating all of the analogy problems from a given batch into a long sequence. In one condition, \textit{misaligned context normalization}, we divided this sequence into segments of length $5$ (as opposed to segments of length $4$ required for context normalization that is aligned with the structure of the task), and computed normalization statistics over these segments. Thus, normalization parameters were computed over segments that intermixed terms (in varying proportion) from two separate analogy problems. In a second control condition, \textit{sliding-window context normalization}, we used a sliding window to compute normalization statistics for each object based on itself and the preceding $3$ objects. Thus, for every object except the last object in each analogy, normalization statistics were computed over a window that incorporated objects from two analogy problems. 

We also compared to a model that employed dropout, a technique proposed to improve generalization in neural networks, in which a random subset of units are dropped from each batch during training \cite{srivastava2014dropout}. Specifically, we implemented a model that combined both batch normalization and 50\% dropout (after normalization), both applied to the output of the feedforward encoder ($\textbf{z}$).

Finally, we compared to a network that did not have any form of normalization applied to it. We found that we had to use a lower learning rate ($1e^-4$) and train for significantly longer ($500,000$ iterations) to achieve convergence with this approach.

\subsection{Dynamic Object Prediction}
\label{dynamic_hyperparams}

For the dynamic object prediction model, we first trained an autoencoder to generate a low-dimensional embedding $\textbf{z}$ given an image $\textbf{x}$. The encoder architecture consisted of $3$ convolutional layers, each with $32$ kernels of size $4\times4$, and a stride of $2$, resulting in a feature map of size $8\times8\times32$. This was followed by $2$ fully-connected layers, with $256$ units per layer. ReLU nonlinearities were used in all layers of the encoder. The image embedding $\textbf{z}$ was then generated with a linear layer consisting of 10 units. 

\begin{table*}[t]
\caption{Results on the Translation Extrapolation regime of the VAEC dataset. Results show accuracy in each region, including the training region (Region 1), averaged over $8$ trained networks ($\pm$ the standard error of the mean).}
\label{extrap_trans_results}
\vskip 0.15in
\begin{center}
\begin{small}
\begin{sc}
\begin{tabular}{lcccccc}
\toprule 
                          & Region 1 (training) & Region 2               & Region 3               & Region 4               & Region 5               & Region 6               \\
Context norm.             & 99.1 $\pm$ 0.6       & \textbf{77.0 $\pm$ 5.8} & \textbf{73.2 $\pm$ 6.4} & \textbf{72.5 $\pm$ 5.1} & \textbf{71.7 $\pm$ 5.5} & \textbf{61.6 $\pm$ 4.9} \\
Sub-batch norm.           & 83.4 $\pm$ 3.3       & 56.1 $\pm$ 1.5          & 51.7 $\pm$ 1.7          & 50.5 $\pm$ 2.2          & 47.0 $\pm$ 2.7          & 46.3 $\pm$ 1.5          \\
Sliding context norm.     & 68.2 $\pm$ 8.2       & 44.8 $\pm$ 1.6          & 35.4 $\pm$ 2.6          & 34.8 $\pm$ 2.7          & 36.7 $\pm$ 4.0          & 36.6 $\pm$ 4.1          \\
Batch + context norm.     & 98.0 $\pm$ 0.5       & 42.5 $\pm$ 1.4          & 33.6 $\pm$ 5.0          & 37.0 $\pm$ 6.3          & 36.7 $\pm$ 6.5          & 36.5 $\pm$ 6.2          \\
Layer norm. (recurrent)   & 100.0 $\pm$ 0.0      & 52.1 $\pm$ 8.0          & 44.5 $\pm$ 7.5          & 35.7 $\pm$ 6.1          & 27.4 $\pm$ 3.9          & 23.3 $\pm$ 3.7          \\
Misaligned context norm.  & 55.3 $\pm$ 4.9       & 41.5 $\pm$ 2.0          & 35.4 $\pm$ 1.0          & 33.3 $\pm$ 1.6          & 31.6 $\pm$ 1.6          & 31.6 $\pm$ 1.8          \\
Batch norm.               & 99.5 $\pm$ 0.1       & 29.0 $\pm$ 2.2          & 26.8 $\pm$ 2.4          & 28.0 $\pm$ 2.6          & 29.0 $\pm$ 2.7          & 30.8 $\pm$ 1.7          \\
Batch norm. + dropout     & 99.0 $\pm$ 0.1       & 27.5 $\pm$ 2.4          & 22.6 $\pm$ 1.8          & 24.4 $\pm$ 2.3          & 26.0 $\pm$ 1.6          & 26.8 $\pm$ 1.7          \\
Layer norm.               & 99.0 $\pm$ 0.2       & 25.2 $\pm$ 2.6          & 22.9 $\pm$ 2.3          & 22.4 $\pm$ 2.3          & 21.6 $\pm$ 2.2          & 17.2 $\pm$ 1.9          \\
No norm.                  & 94.8 $\pm$ 3.3       & 23.9 $\pm$ 2.6          & 23.4 $\pm$ 2.4          & 19.1 $\pm$ 1.9          & 17.8 $\pm$ 1.7          & 16.7 $\pm$ 1.4          \\
Batch norm. (train stats) & 99.9 $\pm$ 0.0       & 33.5 $\pm$ 4.0          & 19.1 $\pm$ 6.4          & 10.2 $\pm$ 4.9          & 13.5 $\pm$ 7.7          & 6.3 $\pm$ 4.3           \\
\bottomrule
\end{tabular}
\end{sc}
\end{small}
\end{center}
\vskip -0.1in
\end{table*}

The decoder architecture consisted of $2$ fully-connected layers with $256$ units, followed by a fully-connected layer with $2,048$ units (reshaped for input to convolutional layers). This was followed by $2$ layers of transposed convolutions, each with $32$ kernels of size $4\times4$, and a fractional stride of $1/2$, and a final transposed convolutional layer with $1$ output channel (also with kernel size $4\times4$ and a fractional stride of $1/2$). ReLU nonlinearities were used in all layers of the decoder, except for the output layer, which used a sigmoid nonlinearity to generate grayscale images with values ranging between $0$ and $1$.

We trained the autoencoder using a reconstruction loss (mean-squared error), with a batch size of $32$, for $200,000$ iterations (though we found that convergence was achieved after approximately $25,000$ iterations), using the ADAM optimizer and a learning rate of $5e^-4$.

After training the autoencoder, we used the encoder to obtain a sequence of low-dimensional embeddings $\textbf{z}_{1}$\ldots$\textbf{z}_{T}$ for each sequence of images. We trained a recurrent network to predict $\textbf{z}_{t}$ given $\textbf{z}_{1}$\ldots$\textbf{z}_{t-1}$. The recurrent network consisted of an LSTM with $20$ units (we found that using larger LSTMs did not make a difference in this task), and a linear output layer with $10$ units, corresponding to the size of its input (i.e. the embedding layer of the autoencoder). 

We performed context normalization before passing the embeddings to the recurrent network, and then de-normalized the predictions made by the recurrent network. We also compare to versions with batch normalization (computed either online using batches of size $N = 32$ on the test set, or by calculating statistics over a sample of size $N = 500$ from the training set), and to a version without any normalization. 

We trained the recurrent network using the mean-squared error between the predicted embedding $\hat{\textbf{z}}$ and the true embedding $\textbf{z}$ with a batch size of $32$, for $20,000$ iterations, using the ADAM optimizer and a learning rate of $5e^-4$. All simulations for the dynamic object prediction task were performed using PyTorch \cite{paszke2017automatic}. All weights and biases were initialized using a uniform distribution bounded by $1 / \sqrt{n}$, where $n$ is the number of input dimensions for a given layer (default method in PyTorch). We evaluate the combined model (including the encoder, LSTM, and decoder) using the mean-squared error between the predicted image $\hat{\textbf{x}}$ and the true image $\textbf{x}$.

\begin{table*}
\caption{Results on the Scale Extrapolation regime of the VAEC dataset. Results show accuracy at each scale, including the training scale (Scale 1), averaged over $8$ trained networks ($\pm$ the standard error of the mean).}
\label{extrap_scale_results}
\vskip 0.15in
\begin{center}
\begin{small}
\begin{sc}
\begin{tabular}{lcccccc}
\toprule
                          & Scale 1 (training) & Scale 2                & Scale 3                & Scale 4                & Scale 5                & Scale 6                \\
Context norm.             & 98.8 $\pm$ 0.7      & \textbf{77.8 $\pm$ 1.8} & \textbf{61.2 $\pm$ 3.8} & \textbf{54.4 $\pm$ 3.3} & \textbf{51.2 $\pm$ 2.5} & \textbf{48.7 $\pm$ 2.2} \\
Layer norm. (recurrent)   & 100.0 $\pm$ 0.0     & 44.1 $\pm$ 5.0          & 28.1 $\pm$ 2.4          & 23.4 $\pm$ 1.6          & 20.2 $\pm$ 1.2          & 18.3 $\pm$ 0.8          \\
Batch norm. (train stats) & 99.9 $\pm$ 0.0      & 40.2 $\pm$ 1.2          & 25.7 $\pm$ 1.7          & 21.2 $\pm$ 1.2          & 21.3 $\pm$ 0.8          & 19.9 $\pm$ 0.8          \\
Batch norm.               & 99.2 $\pm$ 0.2      & 18.3 $\pm$ 0.3          & 17.7 $\pm$ 0.5          & 18.4 $\pm$ 0.4          & 20.1 $\pm$ 0.8          & 21.2 $\pm$ 1.0          \\
No norm.                  & 94.4 $\pm$ 3.5      & 20.9 $\pm$ 1.0          & 17.6 $\pm$ 0.7          & 17.6 $\pm$ 0.6          & 16.7 $\pm$ 0.7          & 16.7 $\pm$ 0.6          \\
\bottomrule
\end{tabular}
\end{sc}
\end{small}
\end{center}
\vskip -0.1in
\end{table*}

\section{Results}

\subsection{VAEC Dataset}

\subsubsection{Translation Extrapolation Regime}

Table~\ref{extrap_trans_results} shows the results on the Translation Extrapolation regime of the VAEC dataset. In general, we find that performance decreases monotonically as a function of distance from the training domain, although we note that most models struggle even with extrapolation from Region 1 to Region 2. This suggests that the VAEC dataset is indeed a challenging benchmark, and an effective method of evaluating extrapolation in neural networks.

Promisingly, we find that networks trained with context normalization extrapolate considerably better than any of the other techniques that we tested. This is particularly true when compared to networks trained without any normalization at all, but there is also a substantial difference in test accuracy when comparing to established techniques, such as layer and batch normalization, with an overall decrease in test error of $42\%$ relative to the next best method (sub-batch normalization).

Some of the techniques that we tested were designed to better understand context normalization. From the performance of these techniques, we learn a few things. First, from the comparison with sub-batch normalization, we learn that the improvement from context normalization is not due merely to normalizing over a smaller sample. Second, from the comparison with both sliding-window and misaligned context normalization, we learn that it is important for context normalization to be aligned with the temporal structure of the task. Third, we learn that combining context normalization with batch normalization actually results in worse generalization than with context normalization alone.

As expected, we also find that networks trained with batch normalization extrapolate quite poorly when statistics are computed over the training set. This result emphasizes an additional strength of context normalization: that it can be computed online from single test samples, rather than requiring batches during evaluation.

One unexpected result was that sub-batch normalization was the second best performing technique. This was surprising because previous work has found batch normalization works better with larger batch sizes \cite{wu2018group}. We speculate that normalizing over small sub-batches might implicitly regularize the learned representations by introducing a source of noise during training, enabling stronger extrapolation. However, we note that batch normalization actually outperforms sub-batch normalization within the training region, suggesting that normalizing over larger batches is indeed better in the traditional IID generalization regime.

Here, we have focused on the benefits of normalization for generalization, but a common reason for applying normalization techniques to neural networks is to decrease training time. We found in our simulations that context normalization provided a comparable acceleration in training speed to batch normalization (training time courses are presented in the Supplementary Material), demonstrating that it is also useful for this purpose.

We also performed an analysis to better understand how context normalization shaped the representations learned by our networks. We found that context normalization encouraged the learning of representations that mirrored the linear structure of the stimulus space, and that this structure was preserved across the test regions in a manner that supported extrapolation (Supplementary Material).

Finally, we note that, although context normalization does indeed enable a substantial increase in the ability to extrapolate, extrapolation is still far from perfect. Thus, we see the VAEC dataset as a tool to aid in the development of methods with even stronger abilities to extrapolate.

\subsubsection{Scale Extrapolation Regime}

Table~\ref{extrap_scale_results} shows the results on the Scale Extrapolation regime. As with the Translation Extrapolation regime, we find that extrapolating between scales is quite challenging, with performance monotonically decreasing as distance from the training domain increases. We find, however, that models trained with context normalization again display a considerable improvement in the ability to extrapolate relative to the other techniques we tested.

\subsection{Visual Analogy Dataset}

Table~\ref{visual_analogy_results} shows the results on the visual analogy dataset from Hill et al. \yrcite{hill2019learning}. We find that applying context normalization over the source and target separately enables a $32\%$ decrease in test error relative to the results of Hill et al. \yrcite{hill2019learning}. We also find that batch normalization, and context normalization applied over the entire analogy problem, both enable more limited improvements in extrapolation. These results show that context normalization can also improve extrapolation in a more complex visual setting.

\begin{table}[t]
\caption{Results for the extrapolation regime of the visual analogy dataset in Hill et al. \yrcite{hill2019learning}. Results show test accuracy averaged over $5$ trained networks for our results, and $10$ trained networks for results from Hill et al. \yrcite{hill2019learning} ($\pm$ the standard error of the mean).}
\label{visual_analogy_results}
\vskip 0.15in
\begin{center}
\begin{small}
\begin{sc}
\begin{tabular}{lc}
\toprule
Context norm. (source/target)   & \textbf{74.2 $\pm$ 0.81}   \\
Context norm. (entire analogy)  & 67.6 $\pm$ 0.49            \\
Batch norm.                     & 66.1 $\pm$ 0.53            \\
Baseline \cite{hill2019learning}& 62 $\pm$ 0.02              \\
\bottomrule
\end{tabular}
\end{sc}
\end{small}
\end{center}
\vskip -0.1in
\end{table}

\subsection{Dynamic Object Prediction}

We find that the generalization benefits of context normalization are not specific to visual analogy problems, but also enable a significant improvement in extrapolation on the dynamic object prediction task. Table~\ref{moving_object_results} shows the average prediction error on the test set (in which objects are, on average, nearly three times the size of objects in the training set), for models trained with context normalization, batch normalization, or no normalization. 

Note that when we implement batch normalization in the conventional manner (computing normalization statistics from the training set), test error is nearly ten times as high as with context normalization. Even when we allow normalization statistics to be computed over the test set, we find that the test error for batch normalization is $70\%$ higher than context normalization. These results demonstrate that the extrapolation benefits afforded by context normalization are not limited to visual analogies, but extend to sequential tasks more generally.

\begin{table}[t]
\caption{Results for the dynamic object prediction task. Results show average MSE for $3$ trained networks on $2$ randomly generated test sets ($\pm$ the standard error of the mean).}
\label{moving_object_results}
\vskip 0.15in
\begin{center}
\begin{small}
\begin{sc}
\begin{tabular}{lc}
\toprule
Context norm.             & \textbf{0.0056 $\pm$ 0.00010}\\
Batch norm.               & 0.0095 $\pm$ 0.00008         \\
Batch norm. (train stats) & 0.0507 $\pm$ 0.00162         \\
No norm.                  & 0.0675 $\pm$ 0.00275         \\
\bottomrule
\end{tabular}
\end{sc}
\end{small}
\end{center}
\vskip -0.1in
\end{table}

\section{Related Work}

Recent studies have investigated the question of extrapolation in neural networks \cite{santoro2018measuring, hill2019learning, saxton2019analysing}. Despite the fact that some of these studies found surprisingly strong performance in complex reasoning tasks, they nevertheless found that current approaches do not perform well when neural networks are required to extrapolate. These results are broadly consistent with the work presented here; however, we note two unique contributions of our work. First, whereas in this previous work neural networks were required to extrapolate to a domain immediately adjacent to the training domain (equivalent to extrapolating from Region $1$ to Region $2$ in our task), the VAEC dataset that we present allows the graded evaluation of extrapolation at distances much farther from the training domain. This is important because, as we find in this work, the ability of neural networks to extrapolate tends to degrade as a function of the distance from the training domain, so the ability to measure extrapolation in terms of this distance is important for evaluating novel approaches to extrapolation. Second, we present a technique that considerably outperforms other approaches at extrapolation, performing reasonably well even in the more challenging evaluation modes of our dataset.

It is important to note that the ability to extrapolate is not the only challenging aspect of analogical reasoning. A related, but separate, challenge arises from the control demands imposed by analogical and relational reasoning tasks more broadly. When many entities are present in a scene or sequence, as is often the case in natural settings, processing the relations between these entities in a systematic manner becomes challenging. A number of architectures have recently been proposed to meet this challenge, with impressive results ranging from relational reasoning \cite{santoro2017simple, santoro2018measuring}, to natural language processing \cite{vaswani2017attention}, to mathematical reasoning \cite{saxton2019analysing}. In the present work, we pursued the hypothesis that the failure of neural networks to extrapolate may be due to the nature of the representations over which they operate, rather than the control demands inherent to relational reasoning tasks. To that end, we focused on a simple problem from a control perspective, allowing us to use a relatively simple recurrent architecture (LSTM). To extend the present approach to more complex settings involving many entities and hierarchical relations, it may be useful to combine our approach with some of these recent architectural developments.

Some studies have employed a `parallelogram' computation ($\hat{D} = C + (B - A)$, based on the approach of Rumelhart et al. \yrcite{rumelhart1973model}) to perform both linguistic \cite{mikolov2013distributed} and visual \cite{reed2015deep} analogies in vector space. Here, we use LSTMs instead of a prespecified computation, with the aim of developing a more flexible framework that is also amenable to other relational and analogical reasoning tasks. The parallelogram approach would not be capable of handling, for instance, either the analogy problems from Hill et al. \yrcite{hill2019learning} or the dynamic object prediction task.

A key aspect of our approach involves normalizing representations with respect to their context. Other forms of normalization have played an important role in recent deep learning research, including batch normalization \cite{ioffe2015batch} and layer normalization \cite{ba2016layer}. These methods have been shown to both speed convergence and improve generalization \cite{bjorck2018understanding}, at least in the traditional IID setting. However, to our knowledge, it has not been tested whether any variants of these methods also improve extrapolation. In our work, we found that normalization can indeed enable a substantial improvement in extrapolation, but the details of the normalization procedure make quite a difference. We found, for instance, that normalizing only over the temporal dimension (`context norm.') results in significantly better extrapolation than normalizing over both the batch and temporal dimensions (`batch + context norm.' in our work, referred to as `sequence-wise normalization' by Laurent et al. \yrcite{laurent2016batch}). 

We note that the idea of normalizing activations with respect to the recent context is similar to the `adaptive detrending' (subtraction of the mean activity over a temporal window) applied by Jung et al. \yrcite{jung2018adaptive}, who found that this procedure improved image recognition from video, providing a benefit both in terms of convergence and (IID) generalization. The details of the normalization procedure in this study were different than ours -- in particular, we implement both a detrending and a scaling operation, as well as the learned gain and shift operations that are commonly found in other normalization procedures. But we are encouraged by the fact that a similar approach proved useful in a more applied setting. Given these results, as well as our results in the dynamic object prediction task, we predict that context normalization may also enable the ability to extrapolate in richer settings such as video prediction.

\section{Discussion}

We have considered the question of how to enable neural networks to extrapolate beyond the convex domain of their training data, making two key contributions. First, we proposed a novel benchmark, the Visual Analogy Extrapolation Challenge (VAEC) dataset, that allows the graded evaluation of extrapolation as a function of distance from the training data, testing for invariance to both scale and translation. Second, we have proposed a simple technique, context normalization, that enables a considerable improvement in the ability to extrapolate, as revealed by experiments using the VAEC dataset, the visual analogy dataset from Hill et al. \yrcite{hill2019learning}, and a dynamic object prediction task.

One possible concern with the benchmark that we propose in this work is that it lacks much of the visual complexity characteristic of real-world data (3D objects, multiple sources of illumination, clutter, etc.). Adding complexity to the VAEC dataset would certainly pose an interesting challenge, but we opted to avoid this in the present work for a simple reason. Adding extraneous complexity to the dataset, unrelated to the issue of extrapolation, would make it difficult to determine whether model failures resulted from this added complexity or from the central challenge of extrapolation. The poor performance on our dataset exhibited by a range of competitive techniques demonstrates that the extrapolation required by the task is more than challenging enough without additional visual complexity. We argue that the VAEC dataset is thus appropriately focused on the highly challenging issue of extrapolation.

Another potential concern with the present work is that context normalization needed to be temporally aligned with the structure of the task in order to enable a significant degree of extrapolation. This was easy to do in the context of our task, but is this too strong a constraint for the approach to be generally applicable? The question of how to align a normalization procedure with the underlying temporal structure of a task is an important one to address in future work, but we highlight two aspects of this problem that are causes for optimism. First, from an engineering perspective, many problems present natural, heuristic methods for segmenting sequential data according to their underlying structure, such as segmenting natural language data at the ends of sentences or paragraphs. Second, we point to a body of work in neuroscience focused precisely on the question of event segmentation \cite{zacks2007event}. In particular, this work suggests particular signatures that might be used as cues to the presence of event boundaries, such as transient changes in prediction error \cite{zacks2011prediction}, or clusters of temporal associations \cite{schapiro2013neural}. 

In this work, we took inspiration from the notion of symmetries in theoretical physics, hypothesizing that data in many domains can be characterized by such symmetries, and that extrapolation can be enabled by designing learning algorithms that exploit these symmetries. Our proposed approach, context normalization, was designed to exploit such symmetries -- in particular, the translation and scale invariance of underlying linear dimensions -- and our results demonstrate that doing so does substantially improve the ability to extrapolate. We emphasize that this result was far from obvious a priori. Though our proposed normalization procedure introduces scale and translational invariance with respect to the representational space (in whichever layer it is applied), this will not necessarily be the same thing as introducing scale and translation invariance with respect to the underlying object space (the size, location, and brightness of objects).

These results are particularly surprising in the case of our experiments with the VAEC dataset and the dataset from Hill et al. \yrcite{hill2019learning}, in which no aspect of the trained system, including the feedforward encoder, experienced any objects outside of the narrowly defined training domain. In this case, the downstream presence of context normalization apparently shaped the learning of representations in the encoder that supported a significant degree of extrapolation. In other words, the learning of representations that support extrapolation was encouraged by the presence of a subtle inductive bias that reflected the underlying symmetries in the task space. Understanding this interaction better is an important task for future work, and would likely lead to even further improvements in the ability to extrapolate.

Finally, we note that there is likely much more to be gained from the design of inductive biases that reflect the underlying symmetries of a given task space. In addition to designing techniques to more effectively capitalize on translation and scale invariance, there are also a host of other symmetries to be exploited, including invariance with respect to rotation in space, translation in time, and so on. There are also a number of opportunities to capitalize on symmetry \textit{between} domains that are characterized by similar underlying structure. This is indeed the basis of more advanced forms of analogical reasoning (e.g., the analogy between the solar system and an atom). It is no coincidence that \textit{complex, relational} representations are the hallmark of analogical reasoning because the most abstract and far-reaching invariances are expressed as systems of relations. We look forward to exploring these ideas in greater depth in future work.

\section*{Acknowledgments}

We would like to thank Timothy Buschman, Simon Segert, Mariano Tepper, Jacob Russin, and the reviewers for helpful feedback and discussions. We would also like to thank David Turner for assistance in performing simulations.

\bibliography{learning_extrapolation}

\begin{thebibliography}{30}
\providecommand{\natexlab}[1]{#1}
\providecommand{\url}[1]{\texttt{#1}}
\expandafter\ifx\csname urlstyle\endcsname\relax
  \providecommand{\doi}[1]{doi: #1}\else
  \providecommand{\doi}{doi: \begingroup \urlstyle{rm}\Url}\fi

\bibitem[Abadi et~al.(2016)Abadi, Barham, Chen, Chen, Davis, Dean, Devin,
  Ghemawat, Irving, Isard, et~al.]{abadi2016tensorflow}
Abadi, M., Barham, P., Chen, J., Chen, Z., Davis, A., Dean, J., Devin, M.,
  Ghemawat, S., Irving, G., Isard, M., et~al.
\newblock Tensorflow: A system for large-scale machine learning.
\newblock In \emph{12th $\{$USENIX$\}$ Symposium on Operating Systems Design
  and Implementation ($\{$OSDI$\}$ 16)}, pp.\  265--283, 2016.

\bibitem[Ba et~al.(2016)Ba, Kiros, and Hinton]{ba2016layer}
Ba, J.~L., Kiros, J.~R., and Hinton, G.~E.
\newblock Layer normalization.
\newblock \emph{arXiv preprint arXiv:1607.06450}, 2016.

\bibitem[Bjorck et~al.(2018)Bjorck, Gomes, Selman, and
  Weinberger]{bjorck2018understanding}
Bjorck, N., Gomes, C.~P., Selman, B., and Weinberger, K.~Q.
\newblock Understanding batch normalization.
\newblock In \emph{Advances in Neural Information Processing Systems}, pp.\
  7694--7705, 2018.

\bibitem[Feynman(1967)]{feynman1967character}
Feynman, R.
\newblock The character of physical law (1965).
\newblock \emph{Cox and Wyman Ltd., London}, 1967.

\bibitem[Feynman(1966)]{feynman1966symmetry}
Feynman, R.~P.
\newblock Symmetry in physical laws.
\newblock \emph{The Physics Teacher}, 4\penalty0 (4):\penalty0 161--174, 1966.

\bibitem[Gentner(1983)]{gentner1983structure}
Gentner, D.
\newblock Structure-mapping: A theoretical framework for analogy.
\newblock \emph{Cognitive science}, 7\penalty0 (2):\penalty0 155--170, 1983.

\bibitem[Glorot \& Bengio(2010)Glorot and Bengio]{glorot2010understanding}
Glorot, X. and Bengio, Y.
\newblock Understanding the difficulty of training deep feedforward neural
  networks.
\newblock In \emph{Proceedings of the thirteenth international conference on
  artificial intelligence and statistics}, pp.\  249--256, 2010.

\bibitem[Hill et~al.(2019)Hill, Santoro, Barrett, Morcos, and
  Lillicrap]{hill2019learning}
Hill, F., Santoro, A., Barrett, D.~G., Morcos, A.~S., and Lillicrap, T.
\newblock Learning to make analogies by contrasting abstract relational
  structure.
\newblock \emph{arXiv preprint arXiv:1902.00120}, 2019.

\bibitem[Holyoak(2012)]{holyoak2012analogy}
Holyoak, K.~J.
\newblock Analogy and relational reasoning.
\newblock \emph{The Oxford handbook of thinking and reasoning}, pp.\  234--259,
  2012.

\bibitem[Ioffe \& Szegedy(2015)Ioffe and Szegedy]{ioffe2015batch}
Ioffe, S. and Szegedy, C.
\newblock Batch normalization: Accelerating deep network training by reducing
  internal covariate shift.
\newblock \emph{arXiv preprint arXiv:1502.03167}, 2015.

\bibitem[Jung et~al.(2018)Jung, Lee, and Tani]{jung2018adaptive}
Jung, M., Lee, H., and Tani, J.
\newblock Adaptive detrending to accelerate convolutional gated recurrent unit
  training for contextual video recognition.
\newblock \emph{Neural Networks}, 105:\penalty0 356--370, 2018.

\bibitem[Kingma \& Ba(2014)Kingma and Ba]{kingma2014adam}
Kingma, D.~P. and Ba, J.
\newblock Adam: A method for stochastic optimization.
\newblock \emph{arXiv preprint arXiv:1412.6980}, 2014.

\bibitem[Lake \& Baroni(2017)Lake and Baroni]{lake2017generalization}
Lake, B.~M. and Baroni, M.
\newblock Generalization without systematicity: On the compositional skills of
  sequence-to-sequence recurrent networks.
\newblock \emph{arXiv preprint arXiv:1711.00350}, 2017.

\bibitem[Lake et~al.(2017)Lake, Ullman, Tenenbaum, and
  Gershman]{lake2017building}
Lake, B.~M., Ullman, T.~D., Tenenbaum, J.~B., and Gershman, S.~J.
\newblock Building machines that learn and think like people.
\newblock \emph{Behavioral and brain sciences}, 40, 2017.

\bibitem[Laurent et~al.(2016)Laurent, Pereyra, Brakel, Zhang, and
  Bengio]{laurent2016batch}
Laurent, C., Pereyra, G., Brakel, P., Zhang, Y., and Bengio, Y.
\newblock Batch normalized recurrent neural networks.
\newblock In \emph{2016 IEEE International Conference on Acoustics, Speech and
  Signal Processing (ICASSP)}, pp.\  2657--2661. IEEE, 2016.

\bibitem[Marcus(2001)]{marcus2001algebraic}
Marcus, G.
\newblock The algebraic mind, 2001.

\bibitem[Mikolov et~al.(2013)Mikolov, Sutskever, Chen, Corrado, and
  Dean]{mikolov2013distributed}
Mikolov, T., Sutskever, I., Chen, K., Corrado, G.~S., and Dean, J.
\newblock Distributed representations of words and phrases and their
  compositionality.
\newblock In \emph{Advances in neural information processing systems}, pp.\
  3111--3119, 2013.

\bibitem[Paszke et~al.(2017)Paszke, Gross, Chintala, Chanan, Yang, DeVito, Lin,
  Desmaison, Antiga, and Lerer]{paszke2017automatic}
Paszke, A., Gross, S., Chintala, S., Chanan, G., Yang, E., DeVito, Z., Lin, Z.,
  Desmaison, A., Antiga, L., and Lerer, A.
\newblock Automatic differentiation in pytorch.
\newblock 2017.

\bibitem[Raven(1941)]{raven1941standardization}
Raven, J.~C.
\newblock Standardization of progressive matrices, 1938.
\newblock \emph{British Journal of Medical Psychology}, 19\penalty0
  (1):\penalty0 137--150, 1941.

\bibitem[Reed et~al.(2015)Reed, Zhang, Zhang, and Lee]{reed2015deep}
Reed, S.~E., Zhang, Y., Zhang, Y., and Lee, H.
\newblock Deep visual analogy-making.
\newblock In \emph{Advances in neural information processing systems}, pp.\
  1252--1260, 2015.

\bibitem[Rumelhart \& Abrahamson(1973)Rumelhart and
  Abrahamson]{rumelhart1973model}
Rumelhart, D.~E. and Abrahamson, A.~A.
\newblock A model for analogical reasoning.
\newblock \emph{Cognitive Psychology}, 5\penalty0 (1):\penalty0 1--28, 1973.

\bibitem[Santoro et~al.(2017)Santoro, Raposo, Barrett, Malinowski, Pascanu,
  Battaglia, and Lillicrap]{santoro2017simple}
Santoro, A., Raposo, D., Barrett, D.~G., Malinowski, M., Pascanu, R.,
  Battaglia, P., and Lillicrap, T.
\newblock A simple neural network module for relational reasoning.
\newblock In \emph{Advances in neural information processing systems}, pp.\
  4967--4976, 2017.

\bibitem[Santoro et~al.(2018)Santoro, Hill, Barrett, Morcos, and
  Lillicrap]{santoro2018measuring}
Santoro, A., Hill, F., Barrett, D., Morcos, A., and Lillicrap, T.
\newblock Measuring abstract reasoning in neural networks.
\newblock In \emph{International Conference on Machine Learning}, pp.\
  4477--4486, 2018.

\bibitem[Saxton et~al.(2019)Saxton, Grefenstette, Hill, and
  Kohli]{saxton2019analysing}
Saxton, D., Grefenstette, E., Hill, F., and Kohli, P.
\newblock Analysing mathematical reasoning abilities of neural models.
\newblock \emph{arXiv preprint arXiv:1904.01557}, 2019.

\bibitem[Schapiro et~al.(2013)Schapiro, Rogers, Cordova, Turk-Browne, and
  Botvinick]{schapiro2013neural}
Schapiro, A.~C., Rogers, T.~T., Cordova, N.~I., Turk-Browne, N.~B., and
  Botvinick, M.~M.
\newblock Neural representations of events arise from temporal community
  structure.
\newblock \emph{Nature neuroscience}, 16\penalty0 (4):\penalty0 486, 2013.

\bibitem[Srivastava et~al.(2014)Srivastava, Hinton, Krizhevsky, Sutskever, and
  Salakhutdinov]{srivastava2014dropout}
Srivastava, N., Hinton, G., Krizhevsky, A., Sutskever, I., and Salakhutdinov,
  R.
\newblock Dropout: a simple way to prevent neural networks from overfitting.
\newblock \emph{The journal of machine learning research}, 15\penalty0
  (1):\penalty0 1929--1958, 2014.

\bibitem[Vaswani et~al.(2017)Vaswani, Shazeer, Parmar, Uszkoreit, Jones, Gomez,
  Kaiser, and Polosukhin]{vaswani2017attention}
Vaswani, A., Shazeer, N., Parmar, N., Uszkoreit, J., Jones, L., Gomez, A.~N.,
  Kaiser, {\L}., and Polosukhin, I.
\newblock Attention is all you need.
\newblock In \emph{Advances in neural information processing systems}, pp.\
  5998--6008, 2017.

\bibitem[Wu \& He(2018)Wu and He]{wu2018group}
Wu, Y. and He, K.
\newblock Group normalization.
\newblock In \emph{Proceedings of the European Conference on Computer Vision
  (ECCV)}, pp.\  3--19, 2018.

\bibitem[Zacks et~al.(2007)Zacks, Speer, Swallow, Braver, and
  Reynolds]{zacks2007event}
Zacks, J.~M., Speer, N.~K., Swallow, K.~M., Braver, T.~S., and Reynolds, J.~R.
\newblock Event perception: a mind-brain perspective.
\newblock \emph{Psychological bulletin}, 133\penalty0 (2):\penalty0 273, 2007.

\bibitem[Zacks et~al.(2011)Zacks, Kurby, Eisenberg, and
  Haroutunian]{zacks2011prediction}
Zacks, J.~M., Kurby, C.~A., Eisenberg, M.~L., and Haroutunian, N.
\newblock Prediction error associated with the perceptual segmentation of
  naturalistic events.
\newblock \emph{Journal of Cognitive Neuroscience}, 23\penalty0 (12):\penalty0
  4057--4066, 2011.

\end{thebibliography}
\bibliographystyle{icml2020}

\end{document}


\twocolumn[
\icmltitle{Supplementary Material}
]

\section{Data and Code Availability}

All code, including code for all simulations and code used to generate the VAEC dataset, is available on \href{https://github.com/taylorwwebb/learning_representations_that_support_extrapolation}{GitHub}. The dataset can be downloaded from \href{https://dataspace.princeton.edu/jspui/handle/88435/dsp01b8515r30h}{DataSpace}.

\section{Architecture and Training Procedure for Visual Analogy Dataset}

For the visual analogy dataset from Hill et al. \yrcite{hill2019learning}, we used an architecture and training procedure modeled as closely as possible on those used in the original paper, so as to facilitate a direct comparison between their model and ours (essentially the original model, with the addition of context normalization).

The model architecture consisted of a feedforward encoder, a recurrent network, and a linear scoring layer. After resizing the images from $160\times160$ to $80\times80$, they were passed to the feedforward encoder, which consisted of $4$ convolutional layers, each with $32$ kernels of size $3\times3$, and a stride of $2$, resulting in a feature map of size $5\times5\times32$. This feature map was then flattened to produce the image embedding. We then applied either context normalization or batch normalization to the embeddings, and passed the embeddings for each candidate analogy (consisting of $3$ source images, $2$ target images, and a candidate answer) to the recurrent network. The recurrent network was a simple RNN, with $64$ hidden units. The final hidden state of the RNN for each candidate analogy was then passed through a linear layer to generate a score for the analogy. 

The scores for all candidate analogies were passed through a softmax layer, and the network was trained to maximize the probability of the correct answer, using a cross entropy loss. We also employ the training method ('Learning Analogies by Contrasting') advocated by Hill et al. \yrcite{hill2019learning} for all simulations with this dataset. Each network was trained with a batch size of $32$ over $3$ full epochs with the entire training set of  $600,000$ analogy problems (we found that all networks reached a clear asymptote by this point in training). We used the ADAM optimizer \cite{kingma2014adam} with a learning rate of $1e^-4$. For the feedforward encoder, all weights were initialized using Kaiming normal initialization \cite{he2015delving}. For the recurrent network and linear scoring layer, weights were initialized using Xavier normal initialization \cite{glorot2010understanding}. All biases were initialized to zero. All simulations for this dataset were performed using PyTorch \cite{paszke2017automatic}.

\begin{table*}
\caption{Analysis of the context normalization model on the Translation Extrapolation regime of the VAEC dataset, separated by individual dimensions of variation. Results show accuracy in each region, including the training region (Region 1), averaged over $8$ trained networks ($\pm$ the standard error of the mean).}
\label{context_norm_results}
\vskip 0.15in
\begin{center}
\begin{small}
\begin{sc}
\begin{tabular}{lcccccc}
\toprule
& Region 1 (training) & Region 2 & Region 3 & Region 4 & Region 5 & Region 6 \\
Size & 97.5 $\pm$ 1.9 & 81.1 $\pm$ 5.3 & 79.8 $\pm$ 8.4 & 79.4 $\pm$ 8.7 & 80.7 $\pm$ 10.0 & 77.0 $\pm$ 11.0 \\
Brightness & 99.0 $\pm$ 0.6 & 97.4 $\pm$ 1.6 & 97.1 $\pm$ 1.2 & 93.4 $\pm$ 2.1 & 90.0 $\pm$ 4.6 & 87.0 $\pm$ 8.0 \\
Location (X) & 99.9 $\pm$ 0.1 & 55.4 $\pm$ 12.2 & 43.9 $\pm$ 9.6 & 47.5 $\pm$ 9.6 & 50.7 $\pm$ 10.2 & 37.3 $\pm$ 7.2 \\
Location (Y) & 100.0 $\pm$ 0.0 & 74.3 $\pm$ 9.7 & 73.2 $\pm$ 11.7 & 64.8 $\pm$ 11.3 & 61.4 $\pm$ 10.5 & 56.5 $\pm$ 9.9 \\
\bottomrule
\end{tabular}
\end{sc}
\end{small}
\end{center}
\vskip -0.1in
\end{table*}

\section{Training Time}

In the work presented here, we have focused on the extent to which our proposed approach, context normalization, improved the ability of neural networks to extrapolate. However, a more common reason for employing normalization techniques in deep learning is to accelerate training \cite{ioffe2015batch, bjorck2018understanding}. We therefore also investigated whether context normalization can provide a similar benefit in terms of training time.

\begin{center}
\centerline{\includegraphics[width=\columnwidth]{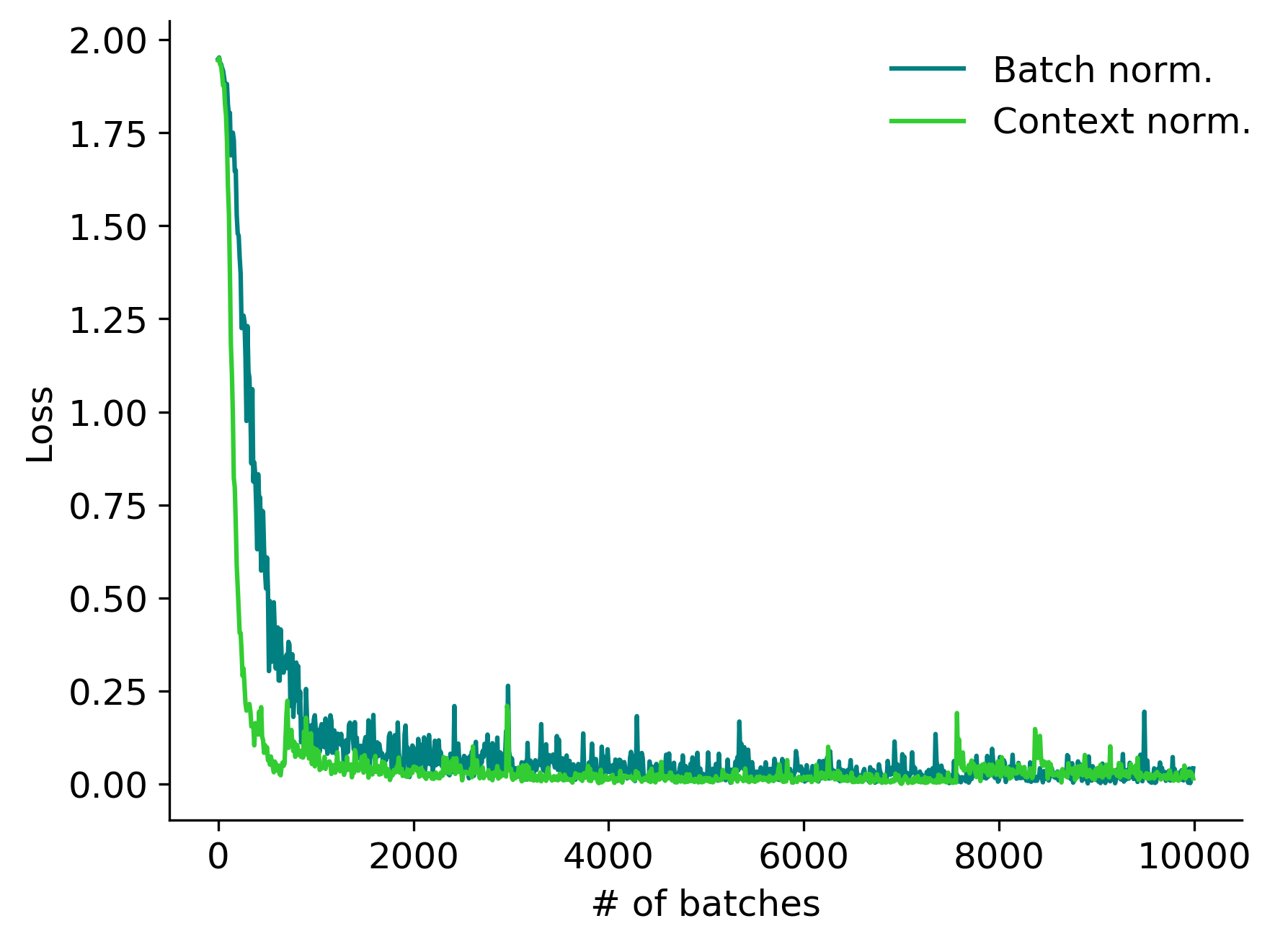}}
\captionof{figure}{Training loss time courses for models trained with context or batch normalization. Each line reflects an average of 8 runs.}
    \label{batch_context_norm}
\end{center}

\begin{center}
\centerline{\includegraphics[width=\columnwidth]{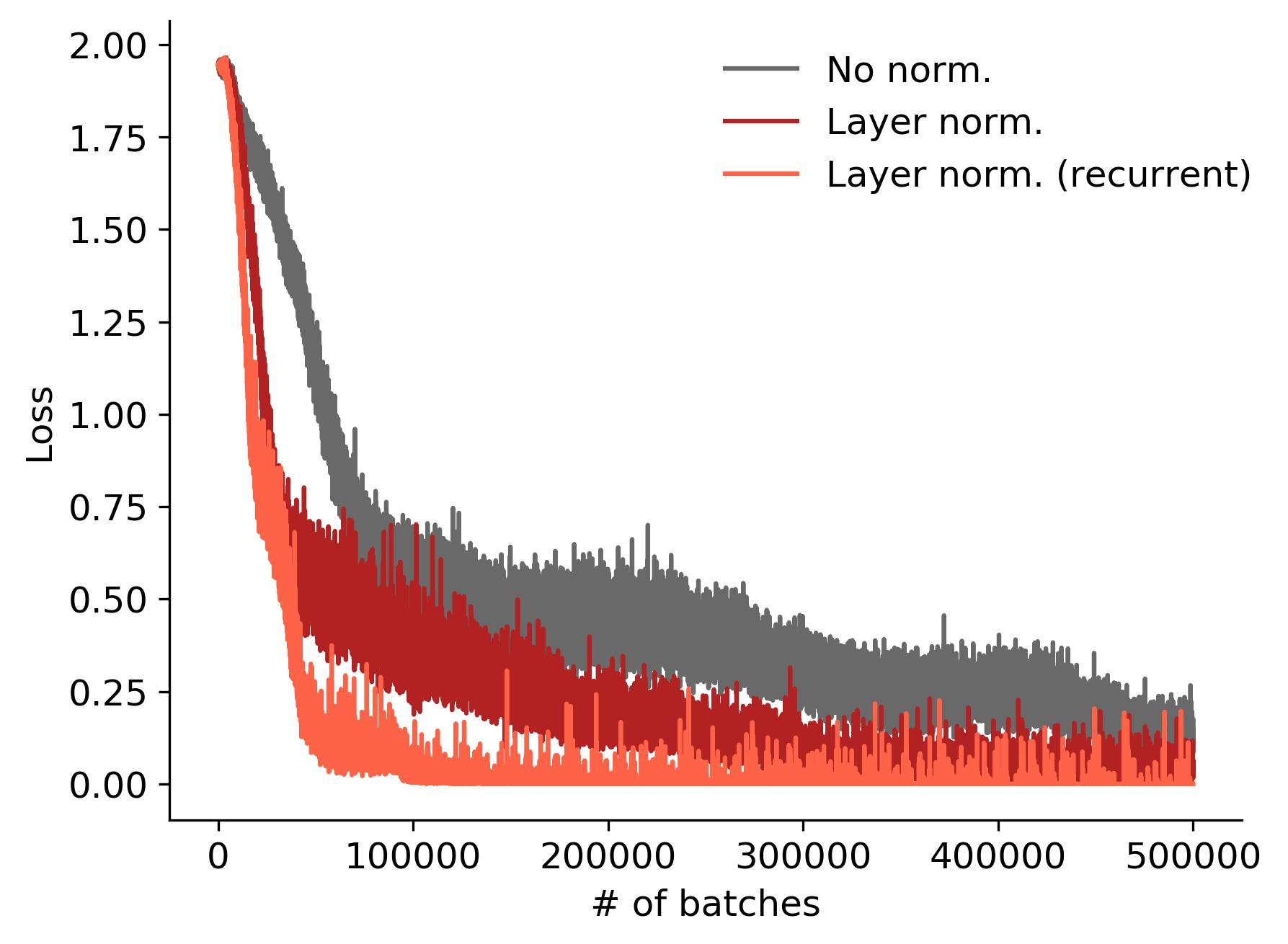}}
\captionof{figure}{Training loss time courses for models trained with layer or no normalization. Each line reflects an average of 8 runs.}
    \label{layer_no_norm}
\end{center}

Figures~\ref{batch_context_norm} and~\ref{layer_no_norm} compare training loss time courses on the Translation Extrapolation regime of the VAEC dataset. Each figure shows training loss averaged across all 8 runs for a particular technique. We find that models trained with both context normalization and batch normalization (Figure~\ref{batch_context_norm}) converge considerably faster than models trained with layer normalization or no normalization at all (Figure~\ref{layer_no_norm}; note the difference in scale of the X axes). We conclude from these results that, in addition to providing a benefit in terms of extrapolation, context normalization also provides a training speedup comparable to that enabled by batch normalization. 

\section{Analysis of Performance Separated by Individual Dimensions of Variation}

The VAEC dataset contains analogy problems in which objects vary in terms of their size, brightness, or location. To determine whether some of these dimensions were more challenging than others, we analyzed performance in each dimension separately. Table~\ref{context_norm_results} shows the results of this analysis. Extrapolation was strongest for the brightness dimension. This may be because this dimension is already represented in the input in a manner that captures the linear structure of the dimension (as a real number). By contrast, our model struggled most with the location dimension. This may be due in part to the convolutional layers in the feedforward encoder, which discard information about location by design. This suggests that one path for improving extrapolation beyond the present results might be to augment the encoder with a scheme for representing spatial location, such as that advocated by Liu et al. \yrcite{liu2018intriguing}.

\section{Learned Representations}

To better understand how context normalization facilitated extrapolation, we also analyzed the latent representations learned by the networks. In our network architecture, each image in a sequence, $\textbf{x}_t$, was processed by a feedforward encoder, yielding a low-dimensional embedding, $\textbf{z}_t$, that was then passed to a recurrent network. We focused our analysis on these learned low-dimensional embeddings.

We performed principal component analysis (PCA) on the learned image embeddings for networks trained on the visual analogy task. PCA was performed separately for the images in each of the training and test regions. For networks trained with context normalization, the embeddings were context-normalized (using the embeddings for the other images in an analogy problem as context) before performing PCA. For networks trained with batch normalization, the embeddings were batch-normalized before performing PCA.

We found that a large amount of variance was captured by the first principal component (PC) for all networks. The first PC accounted for an average of $80\%$ of the variance for networks trained with context normalization, an average of $62\%$ of the variance for networks trained with batch normalization, and an average of $90\%$ of the variance for networks trained with no normalization. Though this may seem surprising, given that the objects in the visual analogy task varied along four dimensions, it can be explained by the fact that, for any given analogy problem, objects only varied along one dimension. Thus, the task can be effectively solved by mapping the objects in an analogy problem to a single dimension, treating the values in irrelevant dimensions as constants.

For each image embedding, we plotted the value of the first PC against the values of the four underlying dimensions of variation. For networks trained with context and batch normalization, we restricted each plot to embeddings from analogy problems in which objects varied along the corresponding dimension (e.g. when plotting the first PC against the size of the object, we restricted the plot to embeddings from analogies in which the size of the object varied). For networks trained without normalization, the learned embeddings did not depend in any way on the other objects in an analogy problem, so this restriction was not applied when plotting the embeddings for networks trained without normalization. 

Additionally, for networks trained with context or batch normalization, we applied the corresponding form of normalization (either context or batch normalization) to the underlying values themselves. We did this because we were primarily interested in the way in which the learned embeddings captured the relative values of the objects, rather than their absolute values. For networks trained without normalization, we used the absolute values, since there was no corresponding normalization procedure to perform.

\begin{figure*}
	\centering \includegraphics[width=\textwidth]{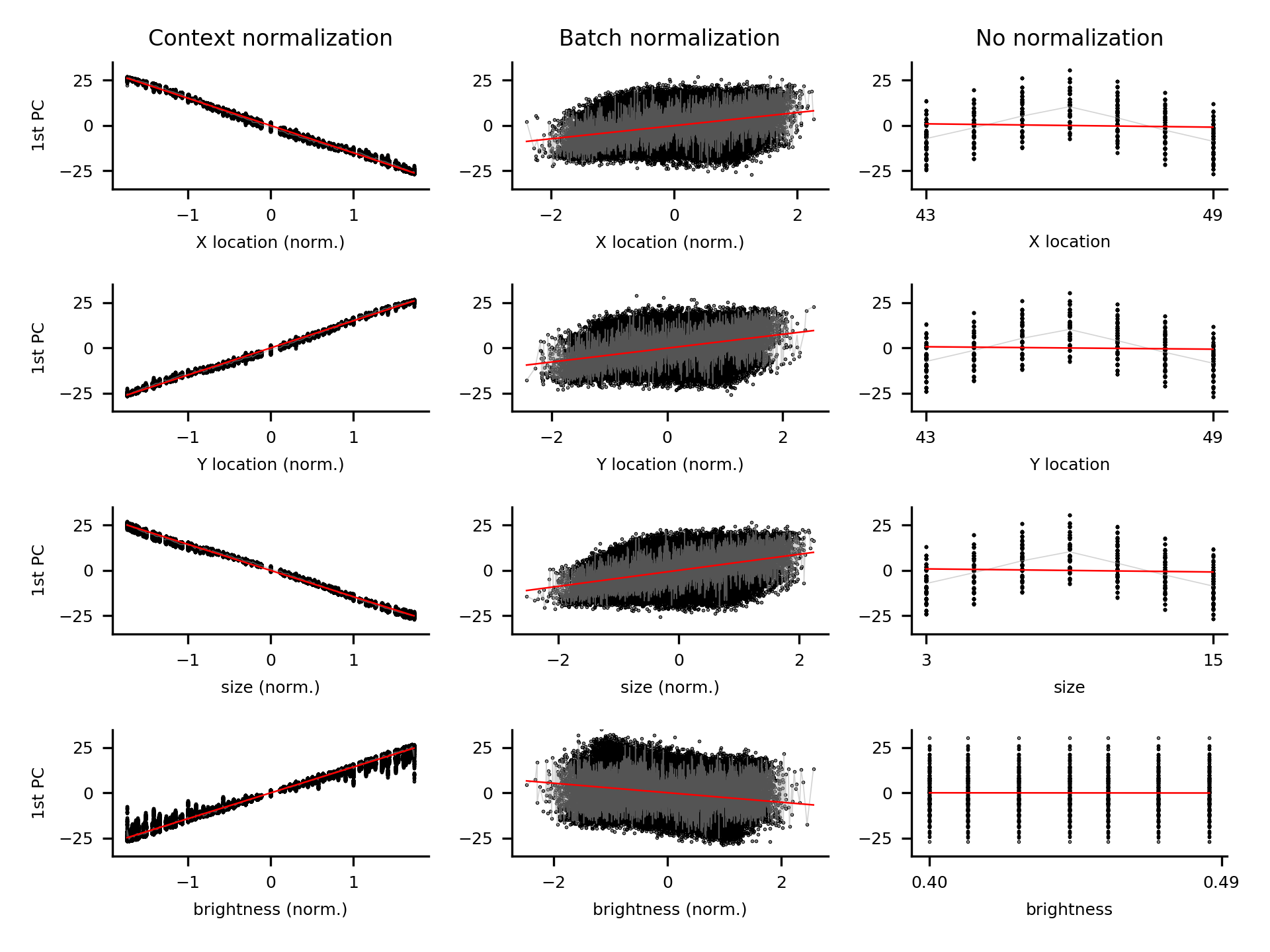}
	\caption{PCA results in Region 1 (training). Y axis represents value along 1st PC. X axis represents value along underlying dimension of variation (location, size, or brightness). Gray line represents the average of all points for each unique value along the X axis. Red line represents a linear fit.}
	\label{train_reps}
\end{figure*}

\begin{figure*}
	\centering \includegraphics[width=\textwidth]{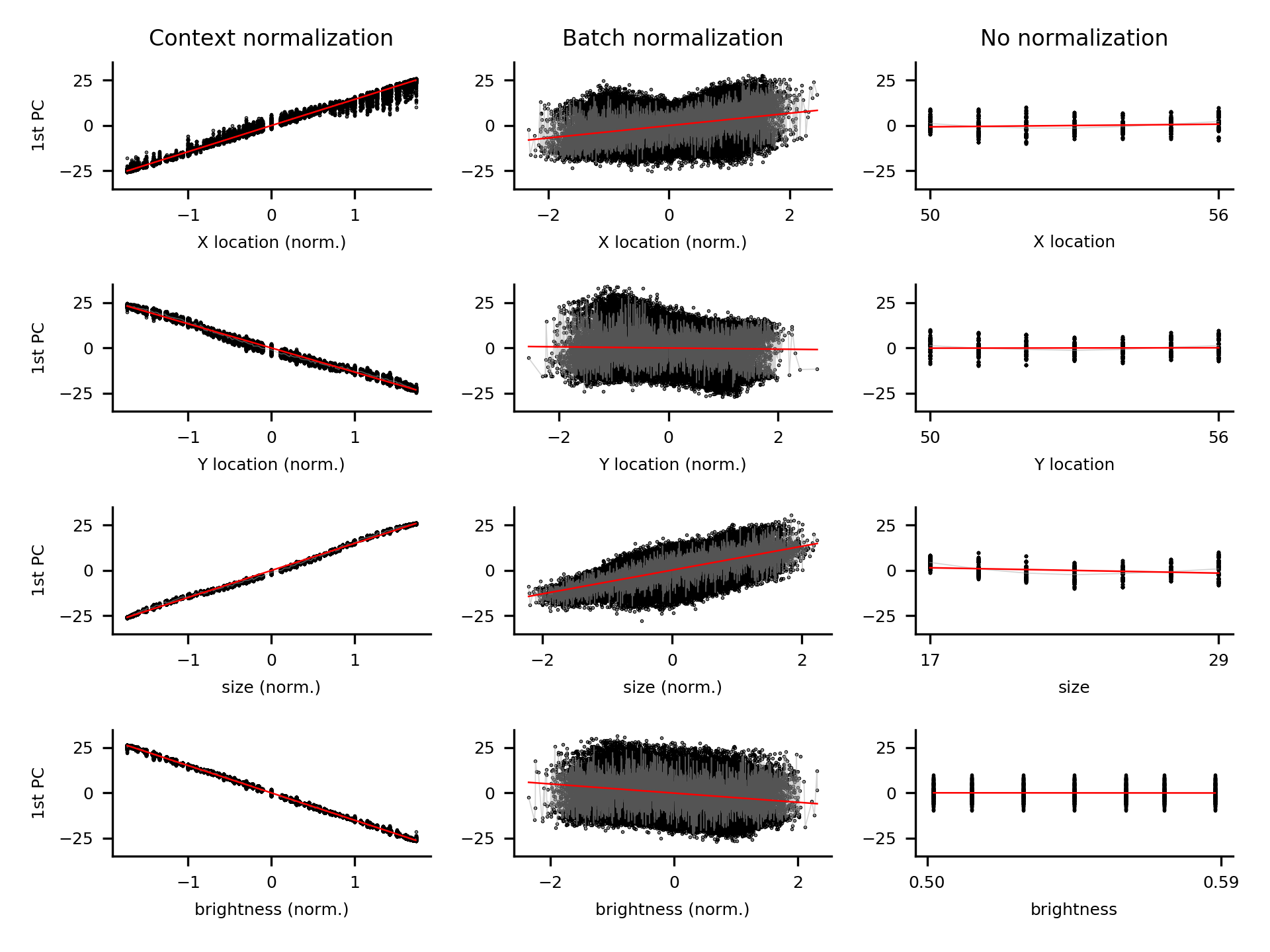}
	\caption{PCA results in Region 2.}
	\label{test1_reps}
\end{figure*}

\begin{figure*}
	\centering \includegraphics[width=\textwidth]{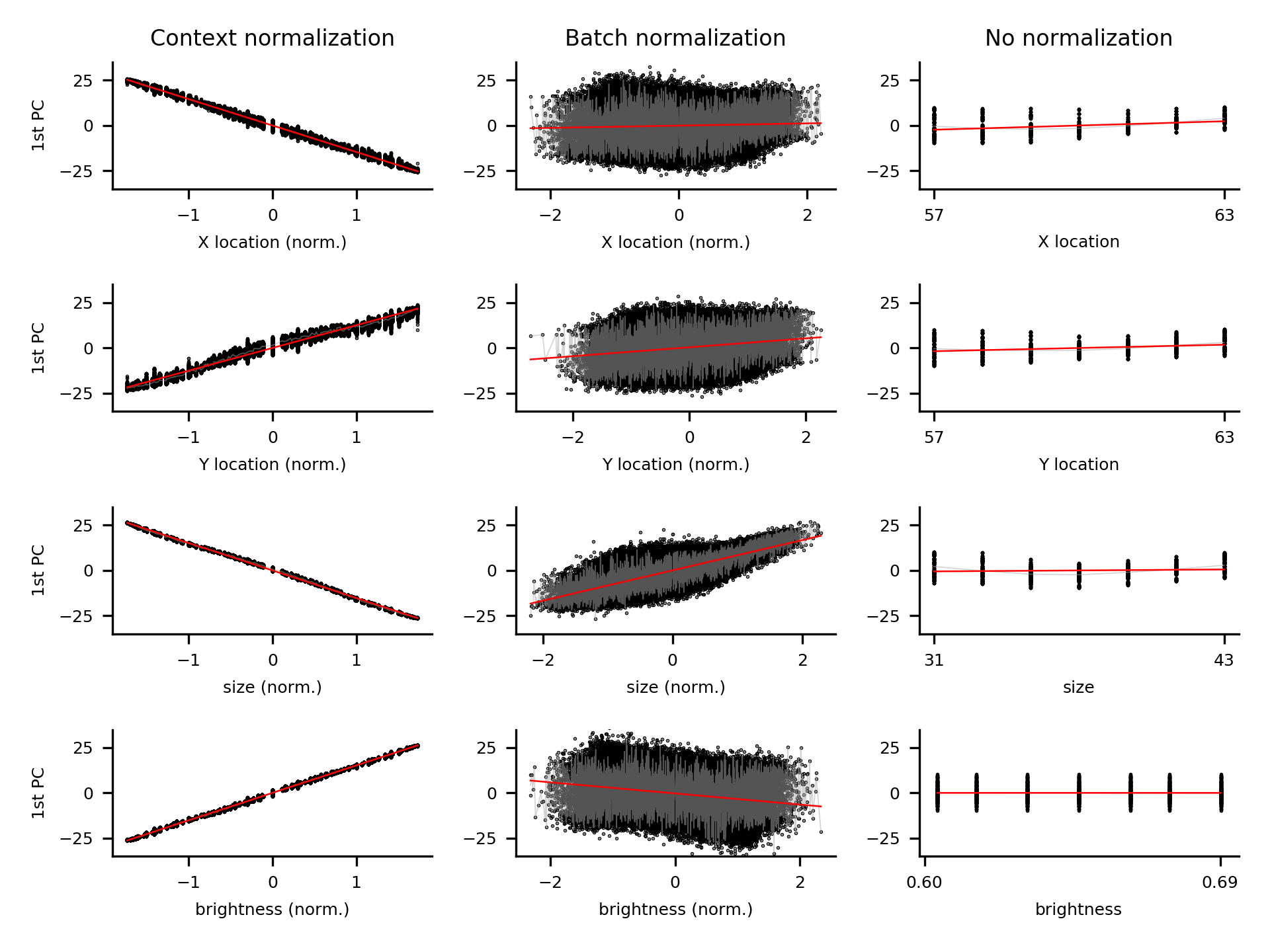}
	\caption{PCA results in Region 3.}
	\label{test2_reps}
\end{figure*}

\begin{figure*}
	\centering \includegraphics[width=\textwidth]{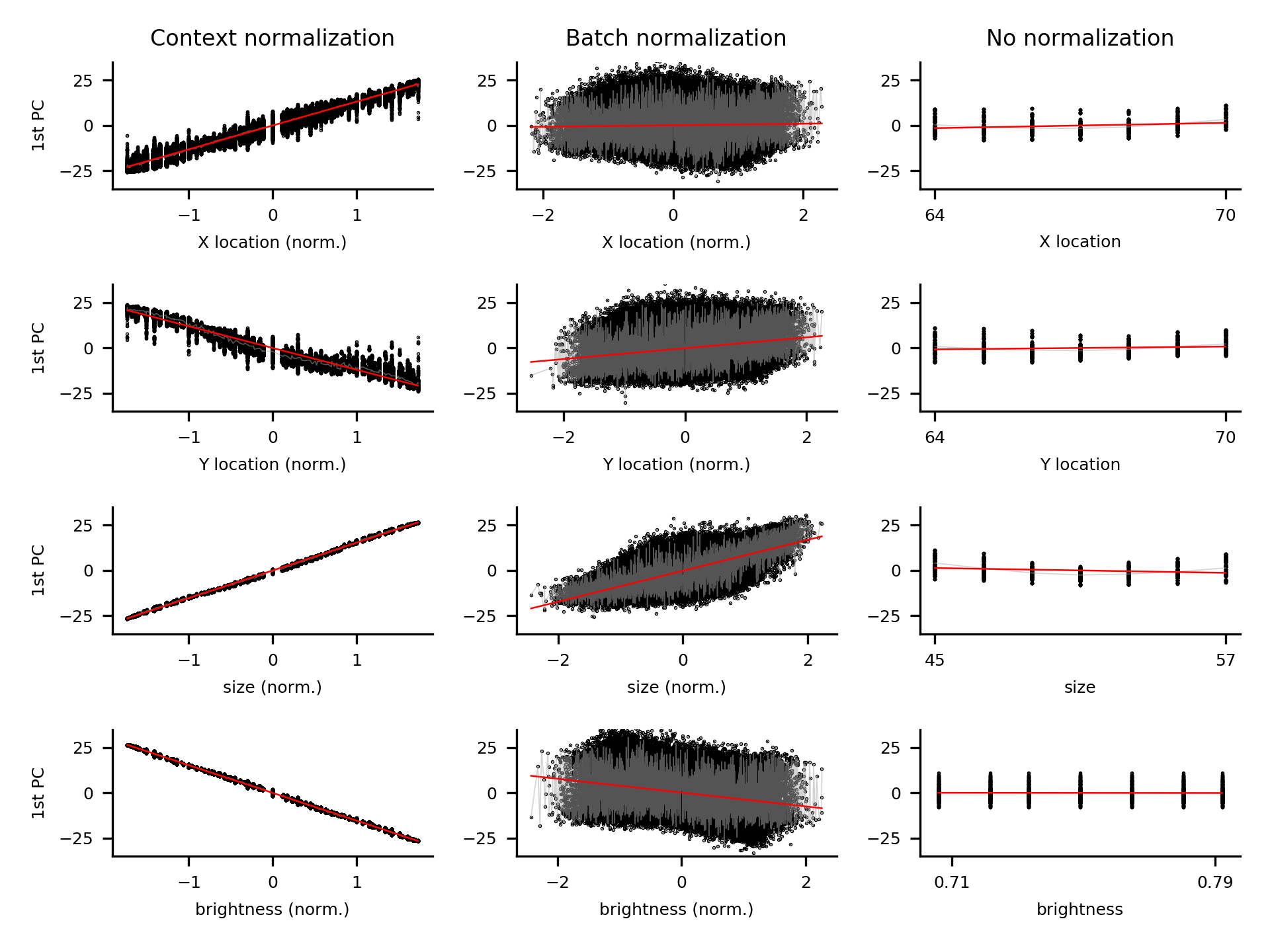}
	\caption{PCA results in Region 4.}
	\label{test3_reps}
\end{figure*}

\begin{figure*}
	\centering \includegraphics[width=\textwidth]{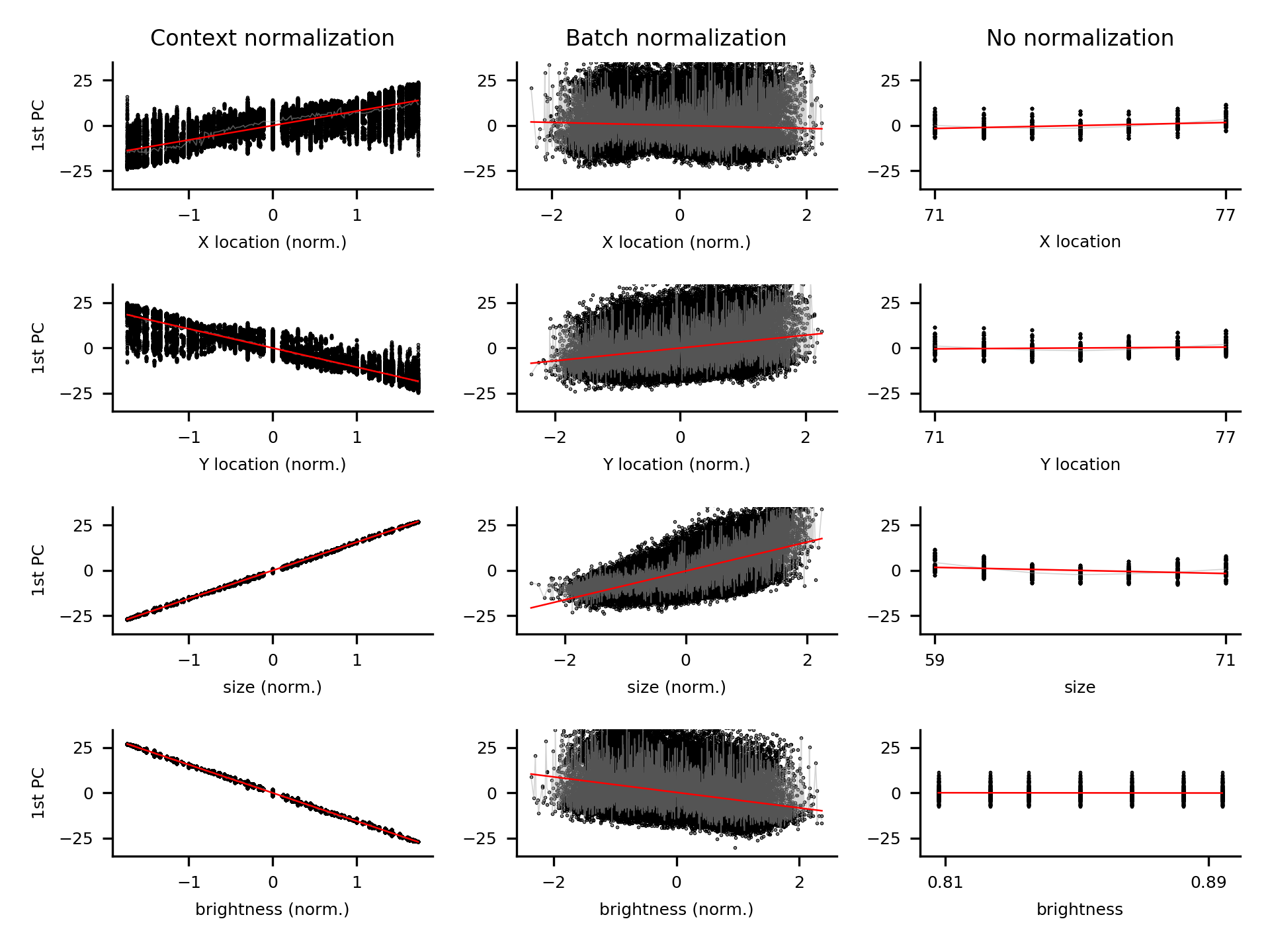}
	\caption{PCA results in Region 5.}
	\label{test4_reps}
\end{figure*}

\begin{figure*}
	\centering \includegraphics[width=\textwidth]{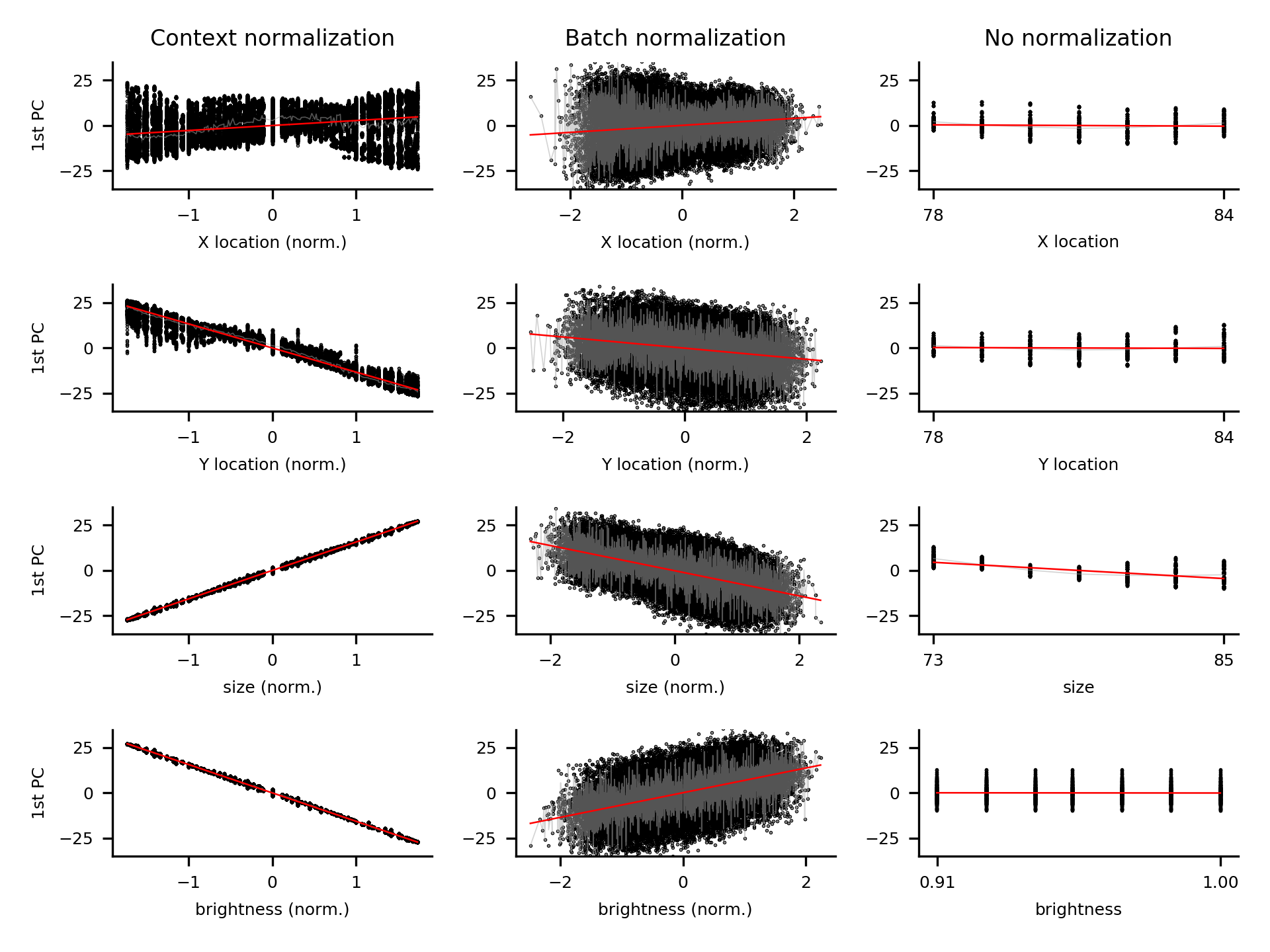}
	\caption{PCA results in Region 6.}
	\label{test5_reps}
\end{figure*}

The results of this analysis are shown in Figures~\ref{train_reps} -~\ref{test5_reps}. The results shown reflect the learned embeddings from the best performing network (as determined by average accuracy in all test regions) with each technique, but the results were qualitatively similar in all networks. In the training region (Figure~~\ref{train_reps}), for networks trained with context normalization, the values along the first PC of the learned embeddings closely tracked a linear function of the underlying values, mirroring the linear structure of the dimensions themselves. By contrast, for networks trained with batch normalization, the relationship between the first PC and underlying values only weakly matched a linear fit. For networks trained without normalization, there was a non-monotonic relationship between the first PC and the underlying values, suggesting that other principal components were likely necessary to accurately represent the values of the objects. 

Moreover, for networks trained with context normalization, the representations learned in the training region showed a considerable degree of similarity to the representations in the test regions (Figures~\ref{test1_reps} -~\ref{test5_reps}; note that the orientation of the linear fits in each region is arbitrary, given that the PCA was performed separately in each region). By contrast, for networks trained with batch normalization or no normalization, there is a greater degree of variability between the representations in each of the training and test regions. For instance, the non-monotonic function observed in the training region for networks trained without normalization is completely absent in the test regions. 

In summary, networks trained with context normalization learned a representation that more closely matched the linear structure of the underlying dimensions, and that was better preserved across the test regions, likely resulting in the improved capacity for extrapolation observed in these networks.

\bibliography{supp_material}
\bibliographystyle{icml2020}